\documentclass[10pt,twocolumn,letterpaper]{article}

\usepackage{wacv}
\usepackage{times}
\usepackage{epsfig}
\usepackage{graphicx}
\usepackage{amsmath}
\usepackage{amssymb}

\usepackage{gensymb}
\usepackage{mathptmx}
\usepackage{multirow}
\usepackage[caption=false]{subfig}
\usepackage[flushleft]{threeparttable}
\usepackage{rotating}
\usepackage[export]{adjustbox}

\usepackage[pagebackref=true,breaklinks=true,letterpaper=true,colorlinks,bookmarks=false]{hyperref}

\wacvfinalcopy 

\def\wacvPaperID{214} 

\ifwacvfinal\fi
\begin{document}

\title{Going Deeper in Facial Expression Recognition using Deep Neural Networks}


\author{Ali Mollahosseini\textsuperscript{1}, David Chan\textsuperscript{2}, and Mohammad H. Mahoor\textsuperscript{1,2}\\
\textsuperscript{1} Department of Electrical and Computer Engineering\\
\textsuperscript{2} Department of Computer Science\\
University of Denver, Denver, CO\\
\tt\small{ali.mollahosseini@du.edu, davidchan@cs.du.edu, and mmahoor@du.edu}
\thanks{This work is partially supported by the NSF grants IIS-1111568 and CNS-1427872.}
\thanks{To be appear in IEEE Winter Conference on Applications of Computer Vision (WACV), 2016}
}

\maketitle

\begin{abstract}
 Automated Facial Expression Recognition (FER) has remained a challenging and interesting problem. Despite efforts made in developing various methods for FER, existing approaches traditionally lack generalizability when applied to unseen images or those that are captured in wild setting. Most of the existing approaches are based on engineered features (e.g. HOG, LBPH, and Gabor) where the classifier's hyperparameters are tuned to give best recognition accuracies across a single database, or a small collection of similar databases. Nevertheless, the results are not significant when they are applied to novel data. This paper proposes a deep neural network architecture to address the FER problem across multiple well-known standard face datasets. Specifically, our network consists of two convolutional layers each followed by max pooling and then four Inception layers. The network is a single component architecture that takes registered facial images as the input and classifies them into either of the six basic or the neutral expressions. We conducted comprehensive experiments on seven publically available facial expression databases, viz. MultiPIE, MMI, CK+, DISFA, FERA, SFEW, and FER2013. The results of proposed architecture are comparable to or better than the state-of-the-art methods and better than traditional convolutional neural networks and in both accuracy and training time.
 
\end{abstract}

\section{Introduction}

Current Human Machine Interaction (HMI) systems have yet to reach the full emotional and social capabilities necessary for rich and robust interaction with human beings. Facial expression, which plays a vital role in social interaction, is one of the most important nonverbal channels through which HMI systems can recognize humans' internal emotions. Ekman \emph{et al.} identified six facial expressions (viz. anger, disgust, fear, happiness, sadness, and surprise) as basic emotional expressions that are universal among human beings~\cite{ekman1971constants}.

Due to the importance of facial expression in designing HMI and Human Robot Interaction (HRI) systems~\cite{mollahosseini2014expressionbot}, numerous computer vision and machine learning algorithms have been proposed for automated Facial Expression Recognition (FER). Also, there exist many annotated face databases with either human actors portraying basic expressions~\cite{gross2010multiPie, pantic2005web, lyons1998coding, mavadati2013disfa}, or faces captured spontaneously in an uncontrolled setting~\cite{dhall2013emotion, mavadati2013disfa}. Automated FER approaches attempt to classify faces in a given single image or sequence of images as one of the six basic emotions. Although, traditional machine learning approaches such as support vector machines, and to a lesser extent, Bayesian classifiers, have been successful when classifying posed facial expressions in a controlled environment, recent studies have shown that these solutions do not have the flexibility to classify images captured in a spontaneous uncontrolled manner (``in the wild'') or when applied databases for which they were not designed~\cite{mayer2014cross}. This poor generalizability of these methods is primarily due to the fact that many approaches are subject or database dependent and only capable of recognizing exaggerated or limited expressions similar to those in the training database. Many FER databases have tightly controlled illumination and pose conditions. In addition, obtaining accurate training data is particularly difficult, especially for emotions such as sadness or fear which are extremely difficult to accurately replicate and do not occur often real life.

Recently, due to an increase in the ready availability of computational power and increasingly large training databases to work with, the machine learning technique of neural networks has seen resurgence in popularity. Recent state of the art results have been obtained using neural networks in the fields of visual object recognition~\cite{krizhevsky2012imagenet, szegedy2014going}, human pose estimation~\cite{toshev2014deeppose}, face verification~\cite{taigman2014deepface}, and many more. Even in the FER field results so far have been promising~\cite{kahou2013combining}. Unlike traditional machine learning approaches where features are defined by hand, we often see improvement in visual processing tasks when using neural networks because of the network's ability to extract undefined features from the training database. It is often the case that neural networks that are trained on large amounts of data are able to extract features generalizing well to scenarios that the network has not been trained on. We explore this idea closely by training our proposed network architecture on a subset of the available training databases, and then performing cross-database experiments which allow us to accurately judge the network's performance in novel scenarios.

In the FER problem, however, unlike visual object databases such as imageNet~\cite{deng2009imagenet}, existing FER databases often have limited numbers of subjects, few sample images or videos per expression, or small variation between sets, making neural networks significantly more difficult to train. For example, the FER2013 database~\cite{FER2013} (one of the largest recently released FER databases) contains 35,887 images of different subjects yet only 547 of the images portray disgust. Similarly, the CMU MultiPIE face database~\cite{gross2010multiPie} contains around 750,000 images but it is comprised of only 337 different subjects, where 348,000 images portray only a ``neutral'' emotion and the remaining images do not portray anger, fear or sadness.

This paper presents a novel deep neural network architecture for the FER problem, and examines the network's ability to perform cross-database classification while training on databases that have limited scope, and are often specialized for a few expressions (e.g. MultiPIE and FERA). We conducted comprehensive experiments on seven well-known facial expression databases (viz. MultiPIE, MMI, CK+, DISFA, FERA, SFEW, and FER2013) and obtain results which are significantly better than, or comparable to, traditional convolutional neural networks or other state-of-the-art methods in both accuracy and learning time.

\section{Background and Related Work}
\label{Sec_BackGround}

Algorithms for automated FER usually involve three main steps, viz. registration, feature extraction, and classification. In the face registration step, faces are first located in the image using some set of landmark points during ``face localization'' or ``face detection''. These detected faces are then geometrically normalized to match some template image in a process called ``face registration''. In the feature extraction step, a numerical feature vector is generated from the resulting registered image. These features can be \textit{geometric features} such as facial landmarks~\cite{kobayashi1997facial}, \textit{appearance features} such as pixel intensities~\cite{Mohammadi2014PCA_based}, Gabor filters~\cite{liu2002gabor}, Local Binary Patterns (LBP)~\cite{shan2009facial}, Local Phase Quantization (LPQ)~\cite{zhen2012facial}, and Histogram of Oriented Gradients (HoG)~\cite{mavadati2013disfa}, or \textit{motion features} such as optical flow~\cite{kenji1991recognition}, Motion History Images (MHI)~\cite{valstar2004motion}, and volume LBP~\cite{zhao2007dynamic}. Current state-of-the-art methods, such as those used in Zhang \emph{et al.}~\cite{zhang2015facial, zhang2014ebear} fuse multiple features using multiple kernel learning algorithms. However by using neural networks, we do not have to worry about the feature selection step - as neural networks have the capacity to learn features that statistically allow the network to make correct classifications of the input data. In the third step, of classification, the algorithm attempts to classify the given face as portraying one of the six basic emotions using machine learning techniques.

Ekman \emph{et al.}~\cite{cohn2007observer, ekman1977facial} distinguished two conceptual approaches to studying facial behavior: a ``message-based" approach and a ``sign-based" approach. Message-based approaches categorize facial behaviors as the the meaning of expressions, whereas sign-based approaches describe facial actions/configuration regardless of the action's meaning. The most well-known and widely used sign-based approach is the Facial Action Coding System (FACS)~\cite{ekman1977facial}. FACS describes human facial movements by their appearance on the face using standard facial substructures called Action Units (AUs). Each AU is based on one or a few facial muscles and AUs may occur individually or in combinations. Similarly, FER algorithms can be categorized into both message-based and sign-based approaches. In message-based approaches FER algorithms are trained on databases labeled with the six basic expressions~\cite{de2011facial}, and more recently, embarrassment and contempt~\cite{lucey2010extended}. Unlike message-based algorithms, sign-based algorithms are trained to detect AUs in a given image or sequence of images~\cite{de2011facial}. These detected AUs are then converted to emotion-specified expressions using EMFACS~\cite{friesen1983emfacs} or similar systems~\cite{taheri2014structure}. In this paper, we develop a message-based neural network solution,

FER systems are traditionally evaluated in either a subject independent manner or a cross-database manner. In subject independent evaluation, the classifier is trained on a subset of images in a database (called the training set) and evaluated on faces in the same database that are not elements of the training set often using K-fold cross validation or leave-one-subject-out approaches. The cross-database method of evaluating facial expression systems requires training the classifier on all of the images in a single database and evaluating the classifier on a different database which the classifier has never seen images from. As single databases have similar settings (illumination, pose, resolution etc.), subject independent tasks are easier to solve than cross database tasks. Subject independent evaluation is not, however, unimportant. If a researcher can guarantee that the data will align well in pose, illumination and other factors with the training set, subject independent evaluation can give a reasonably good representation of the classification accuracy in an online system. Another technique, subject dependent evaluation (person-specific), is also used in limited cases, e.g. FERA 2011 challenge~\cite{valstar2011first}; often in these scenarios the recognition accuracy is more important than the generalization.

Recent approaches to visual object recognition tasks, and the FER problem have used increasingly ``deep'' neural networks (neural networks with large numbers of hidden layers). The term ``deep neural network'' refers to a relatively new set of techniques in neural network architecture design that were developed in order to improve the ability of neural networks to tackle big-data problems. With the large amount of available computing power continuing to grow, deep neural network architectures provide a learning architecture based in the development of ``brain-like'' structures which can learn multiple levels of representation and abstraction which allow algorithms for finding complex patterns in images, sound, and text.

It seems only logical to extend cutting-edge techniques in the field of ``deep learning'' to the FER problem. Deep networks have a remarkable ability to perform well in flexible learning tasks, such as the cross-database evaluation situation, where it is unlikely that hand-crafted features will easily generalize to a new scenario. By training neural networks, particularly deep neural networks, for feature recognition and extraction we can drastically reduce the amount of time that is necessary to implement a solution to the FER problem that, even when confronted with a novel data source, will be able to perform at high levels of accuracy. Similarly, we see deep neural networks performing well in the subject independent evaluation scenarios, as the algorithms can learn to recognize subtle features that even field experts can miss. These correlations provide the motivation for this paper, as the strengths of deep learning seem to align perfectly with the techniques required for solving difficult ``in the wild'' FER problems.

A subset of deep neural network architectures called ``convolutional neural networks'' (CNNs) have become the traditional approach for researchers studying vision and deep learning. In the 2014 ImageNet challenge for object recognition, the top three finishers all used a CNN approach, with the GoogLeNet architecture achieving a remarkable 6.66\% error rate in classification~\cite{szegedy2014going, russakovsky2014imagenet}. The GoogLeNet architecture uses a novel multi-scale approach by using multiple classifier structures, combined with multiple sources for back propagation. This architecture defeats a number of problems that occur when back-propagation decays before reaching beginning layers in the architecture. Additional layers that reduce dimension allow GoogLeNet to increase in both width and depth without significant penalties, and take an elegant step towards complicated network-in-network architectures described originally in Lin \emph{et al.}~\cite{lin2013network}. In other word, the architecture is composed of multiple ``Inception" layers, each of which acts like a micro-network in the larger network, allowing the architecture to make more complex decisions.

More traditional CNN architectures have also achieved remarkable results. AlexNet~\cite{krizhevsky2012imagenet} is an architecture that is based on the traditional CNN layered architecture - stacks of convolutions layers followed by max-pooling layers and rectified linear units (ReLUs), with a number of fully connected layers at the top of the layer stack. Their top=5 error rate of 15.3\% on the ILSVRC-2012 competition revolutionized the way that we think about the effectiveness of CNNs. This network was also one of the first networks to introduce the ``dropout'' method for solving the over fitting problem (Suggested by Hinton \emph{et al.}~\cite{srivastava2014dropout}) which proved key in developing large neural networks. One of the large challenges to overcome in the use of traditional CNN architectures is their depth and computational complexity. The full AlexNet network performs on the order of 100M operations for a single iteration, while SVM and shallow neural networks perform far fewer operations in order to create a suitable model. This makes traditional CNNs very hard to apply in time restrictive scenarios.

In~\cite{liu2013aware} a new deep neural network architecture, called an ``AU-Aware'' architecture was proposed in order to investigate the FER problem. In an AU-Aware architecture, the bottom of the layer stack consists of convolution layers and max-pooling layers which are used to generate a complete representation of the face. Next in the layer stack, an ``AU-aware receptive field layer'' generates a complete representation over all possible spatial regions by convolving the dense-sampling facial patches with special filters in a greedy manner. Then, a multilayer Restricted Boltzmann Machine (RBM) is exploited to learn hierarchical features. Finally, the outputs of the network are concatenated as features which are used to train a linear SVM classifier for recognizing the six basic expressions. Results in~\cite{liu2013aware} show that the features generated by this ``AU-Aware'' network are competitive with or superior to handcrafted features such as LBP, SIFT, HoG, and Gabor on the CK+, MMI and databases using a similar SVM. However, AU-aware layers do not necessarily detect FACS defined action units in faces.

In~\cite{kahou2013combining} multiple deep neural network architectures are combined to solve the FER problem in video analysis. These network architectures included: (1) an architecture similar to the AlexNet CNN run on individual frames of the video, (2) a deep belief network trained on audio information, (3) an autoencoder to model the spatiotemporal properties of human activity, and (4) a shallow network focused on the mouth. The CNN is trained on the private Toronto Face Database~\cite{susskind2010toronto} and fine tuned on the AFEW database~\cite{dhall2013emotion}, yielded an accuracy of 35.58\% when evaluated in a subject independent manner on AFEW. When combined with a single predictor, the five architectures produced an accuracy of 41.03\% on the test set, the highest accuracy in the EmotiW 2013~\cite{dhall2013emotion} challenge, where challenge winner 2014~\cite{liu2014combining} achieved 50.40\% on test set using multiple kernel methods on Riemannian manifold.   

A 3D CNN with deformable action parts constraints is introduced in~\cite{liu2014deeply} which can detect specific facial action parts
under the structured spatial constraints, and obtain the discriminative part-based representation simultaneously. The results on two posed expression datasets, CK+, MMI, and a spontaneous dataset FERA achieve state-of-the-art video-based expression recognition accuracy.

\section{Proposed Method}

Often improving neural network architectures has relied on increasing the number of neurons or increasing the number of layers, allowing the network to learn more complex functions; however, increasing the depth and complexity of a topology leads to a number of problems such as increased over-fitting of training data, and increased computational needs. A natural solution to the problem of increasingly dense networks is to create deep sparse networks, which has both biological inspiration, and has firm theoretical foundations discussed in Arora \etal~\cite{arora2013provable}. Unfortunately, current GPUs and CPUs do not have the capability to efficiently compute actions on sparse networks. The Inception layer presented in~\cite{russakovsky2014imagenet} attempts to rectify these concerns by providing an approximation of sparse networks to gain the theoretical benefits proposed by Arora \etal, however retains the dense structure required for efficient computation.

Applying the Inception layer to applications of Deep Neural Network has had remarkable results, as implied by~\cite{deepid2015} and~\cite{szegedy2014going}, and it seems only logical to extend state of the art techniques used in object recognition to the FER problem. In addition to merely providing theoretical gains from the sparsity, and thus, relative depth, of the network, the Inception layer also allows for improved recognition of local features, as smaller convolutions are applied locally, while larger convolutions approximate global features. The increased local performance seems to align logically with the way that humans process emotions as well. By looking at local features such as the eyes and mouth, humans can distinguish the majority of the emotions~\cite{emorecogBal2010}. Similarly, children with autism often cannot distinguish emotion properly without being told to remember to look at the same local features~\cite{emorecogBal2010}. By using the Inception layer structure and applying the network-in-network theory proposed by Lin \etal~\cite{lin2013network}, we can expect significant gains on local feature performance, which seems to logically translate to improved FER results.

Another benefit of the network-in-network method is that along with increased local performance, the global pooling performance is increased and therefore it is less prone to overfitting. This resistance to overfitting allows us to increase the depth of the network significantly without worrying about the small corpus of images that we are working with in the FER problem.

\begin{figure}
\centering
\adjincludegraphics[width=4in,trim={4cm 0 0 0},clip]{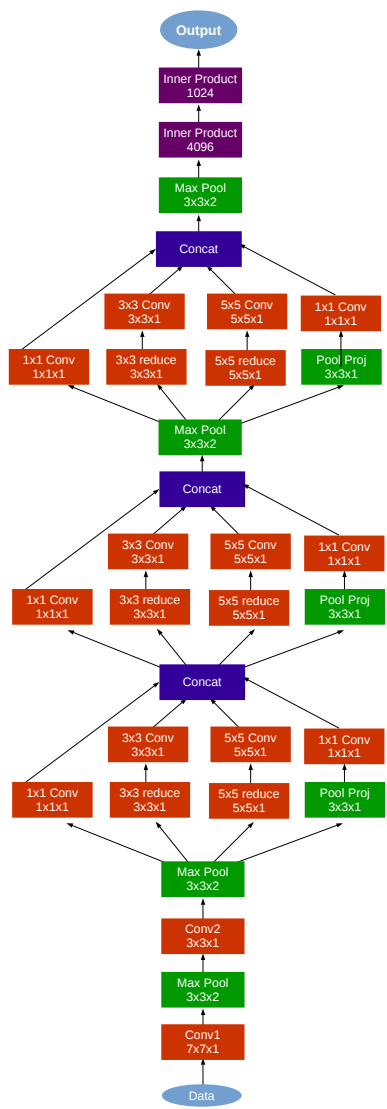}
\vspace{-1.2cm}
\caption{\label{fig:Network_Architecture} Network Architecture}
\end{figure}

The work that we present in this paper is inspired by the techniques provided by the GoogLeNet and AlexNet architectures described in Sec.~\ref{Sec_BackGround}. Our network consists of two elements, first our network contains of two traditional CNN modules (a traditional CNN layer consists of a convolution layer by a max pooling layer). Both of these modules use rectified linear units (ReLU) which have an activation function described by: $$ f(x) = max(0,x) $$ where $x$ is the input to the neuron~\cite{krizhevsky2012imagenet}. Using the ReLU activation function allows us to avoid the vanishing gradient problem caused by some other activation functions (for more details see~\cite{krizhevsky2012imagenet}). Following these modules, we apply the techniques of the network in network architecture and add two ''Inception'' style modules, which are made up of a $1\times 1$, $3\times 3$ and $5\times 5$ convolution layers (Using ReLU) in parallel. These layers are then concatenated as output and we use two fully connected layers as the classifying layers (Also using ReLU). Figure~\ref{fig:Network_Architecture} shows the architecture of the network used in this paper.

\begin{figure}
\centering
\subfloat
{
	\centering
    \includegraphics[width=15mm]{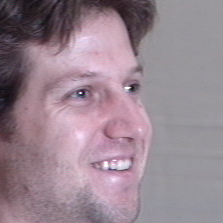}
}
\subfloat
{
	\centering
    \includegraphics[width=15mm]{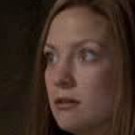}
}
\subfloat
{
	\centering
    \includegraphics[width=15mm]{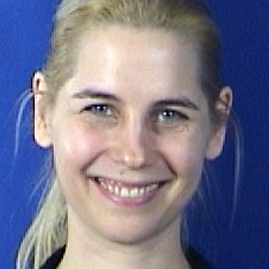}
}
\subfloat
{
	\centering
    \includegraphics[width=15mm]{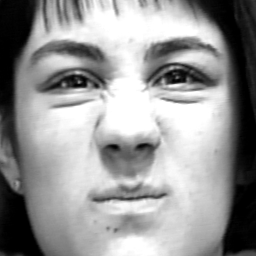}
}
\subfloat
{
	\centering
    \includegraphics[width=15mm]{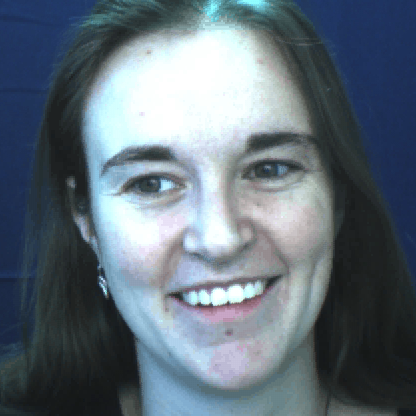}
}
\vspace{-0.2cm}
\subfloat
{
	\centering
    \includegraphics[width=15mm]{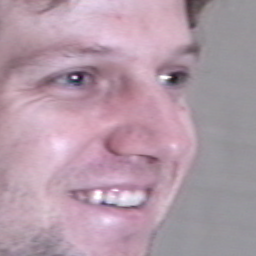}
}
\subfloat
{
	\centering
    \includegraphics[width=15mm]{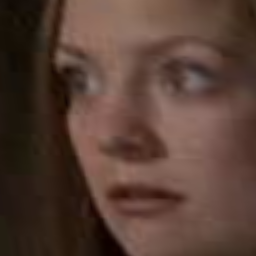}
}
\subfloat
{
	\centering
    \includegraphics[width=15mm]{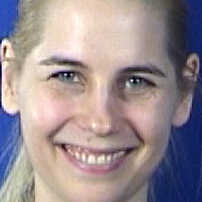}
}
\subfloat
{
	\centering
    \includegraphics[width=15mm]{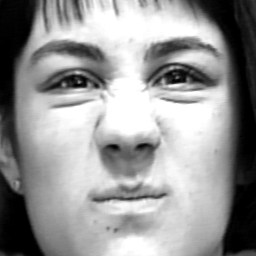}
}
\subfloat
{
	\centering
    \includegraphics[width=15mm]{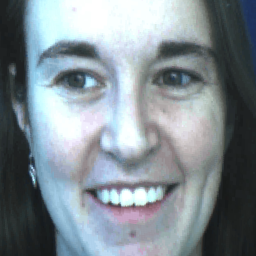}
}
\caption{\label{fig:RegistredImages}
Sample of the face registration. From left to right images are taken from MultiPIE, SFEW, MMI, CK+ and DISFA. First row shows the original images and the second row shows their registered images respectively.
}
\end{figure}

In this work we register facial images in each of the databases using research standard techniques. We used bidirectional warping of Active Appearance Model (AAM)~\cite{mollahosseini2013bidirectional} and a Supervised Descent Method (SDM) called IntraFace~\cite{xiong2013supervised} to extract facial landmarks, however further work could consider improving the landmark recognition in order to extract more accurate faces. IntraFace uses SIFT features for feature mapping and trains a descent method by a linear regression on training set in order to extract 49 points. We use these points to register faces to an average face in an affine transformation. Finally, a fixed rectangle around the average face is considered as the face region. Figure~\ref{fig:RegistredImages} demonstrates samples of the face registration with this method. In our research, facial registration increased the accuracy of our FER algorithms by 4-10\%, which suggests that registration (like normalization in traditional problems) is a significant portion of any FER algorithm.

Once the faces have been registered, the images are resized to 48$\times$48 pixels for analysis. Even though many databases are composed of images with a much higher resolution testing suggested that decreasing this resolution does not greatly impact the accuracy, however vastly increases the speed of the network. To augment our data, we extract 5 crops of 40$\times$40 from the four corners and the center of the image and utilize both of them and their horizontal flips for a total of 10 additional images. 

In training the network, the learning rates are decreased in a polynomial fashion as: $base\_lr(1 - iter/max\_iter)^{0.5}$, where $base\_lr=0.01$ is the base learning rate, $iter$ is the current iteration and $max\_iter$ is the maximum allowed iterations. Testing suggested that other popular learning rate policies such as \textit{fixed} learning rate, \textit{step} where learning rate is multiplies by a gamma factor in each step, and exponential approach did not perform as well as the polynomial fashion. Using the polynomial learning rate, the test loss converged faster and allowed us to train the network for many iterations without the need for fine-tuning. We also trained the bias nodes twice as fast as the weights of the network, in order to increase the rate at which unnecessary nodes are removed from evaluation. This decreases the number of iterations that the network must run before the loss converges.

\begin{table*}[t]
\caption{Network Configuration}
\label{tbl_netCongiguration}
\centering
\begin{tabular}{|l|l|l|l|l|l|l|l|l|l|}
    \hline
    \textbf{\scriptsize{Layer type}} & \textbf{\scriptsize{Patch Size / Stride}} & \textbf{\scriptsize{Output}}                   & \textbf{\scriptsize{1 x 1}} & \textbf{\scriptsize{3 x 3}} & \textbf{\begin{tabular}[c]{@{}l@{}}\scriptsize{3 x 3 reduce}\end{tabular}} & \textbf{\scriptsize{5 x 5}} & \textbf{\begin{tabular}[c]{@{}l@{}}\scriptsize{5 x 5 reduce}\end{tabular}} & \textbf{\scriptsize{Proj Pooling}} & \textbf{\scriptsize{\# Operations}} \\ \hline
    \scriptsize{Convolution - 1}                & \scriptsize{$7 \times 7$ / 2 }            & \scriptsize{$24\times24\times64$ }    & ~                                                 & ~                                                 & ~                                                                                                                            & ~                                                 & ~                                                                                                                            & ~                                                        & \scriptsize{5.7M}                                                \\ \hline
    \scriptsize{Max pool - 1}                   & \scriptsize{$3 \times 3$} / 2             & \scriptsize{$12\times12\times64$ }    & ~                                                 & ~                                                 & ~                                                                                                                            & ~                                                 & ~                                                                                                                            & ~                                                        & \scriptsize{5.7M}                                                \\ \hline
    \scriptsize{Convolution - 2}                & \scriptsize{$3\times3$} / 1               & \scriptsize{$12\times12\times192$ }   & ~                                                 & ~                                                 & ~                                                                                                                            & ~                                                 & ~                                                                                                                            & ~                                                        & \scriptsize{1.4M}                                                \\ \hline
    \scriptsize{Max Pool - 2 }                  & \scriptsize{$3 \times3$ / 2     }         & \scriptsize{$6 \times 6 \times 192$ } & ~                                                 & ~                                                 & ~                                                                                                                            & ~                                                 & ~                                                                                                                            & ~                                                        & \scriptsize{1.4M}                                                \\ \hline
    \scriptsize{Inception - 3a }                & ~                                                               & ~                                                                    & \scriptsize{64    }                    & \scriptsize{128    }                   & \scriptsize{96}                                                                                                   & \scriptsize{32              }          & \scriptsize{16  }                                                                                                 & \scriptsize{32}                               & \scriptsize{2.6M}                                                \\ \hline
    \scriptsize{Inception - 3b }                & ~                                                               & ~                                                                    & \scriptsize{128      }                 & \scriptsize{192   }                    & \scriptsize{128                                                              }                                    & \scriptsize{96      }                  & \scriptsize{32                                                               }                                    & \scriptsize{64    }                           & \scriptsize{4.5M}                                                \\ \hline
    \scriptsize{Max Pool - 4}                   & \scriptsize{$3 \times 3$ / 2  }           & \scriptsize{$3 \times 3 \times 480$ } & ~                                                 & ~                                                 & ~                                                                                                                            & ~                                                 & ~                                                                                                                            & ~                                                        & \scriptsize{0.6M}                                                \\ \hline
    \scriptsize{Inception - 4a }                & ~                                                               & ~                                                                    & \scriptsize{192   }                    & \scriptsize{208 }                      & \scriptsize{96                                                               }                                    & \scriptsize{48     }                   & \scriptsize{16                                                               }                                    & \scriptsize{64   }                            & \scriptsize{1.3M}                                                \\ \hline
    \scriptsize{Avg Pooling - 6}             ~ & ~                                                               & \scriptsize{$1 \times 1 \times 1024$} & ~                                                 & ~                                                 & ~                                                                                                                            & ~                                                 & ~                                                                                                                            & ~                                                        & \scriptsize{25.6K}                                                \\ \hline
    \scriptsize{Fully Connected - 7}            & ~                                                               & \scriptsize{$1 \times 1 \times 4096$} & ~                                                 & ~                                                 & ~                                                                                                                            & ~                                                 & ~                                                                                                                            & ~                                                        & \scriptsize{0.2M}                                                \\ \hline
    \scriptsize{Fully Connected - 8}            & ~                                                               & \scriptsize{$1 \times 1 \times 1024$} & ~                                                 & ~                                                 & ~                                                                                                                            & ~                                                 & ~                                                                                                                            & ~                                                        & \scriptsize{51K}                                                \\ \hline
    \end{tabular}
\end{table*}

\section{Experimental Results}

\subsection{Face Databases}
We evaluate the proposed method on well-known publicly available facial expression databases: CMU MultiPIE~\cite{gross2010multiPie}, MMI~\cite{pantic2005web}, Denver Intensity of Spontaneous Facial Actions (DISFA)~\cite{mavadati2013disfa}, extended CK+~\cite{lucey2010extended}, GEMEP-FERA database~\cite{banziger2010introducing}, SFEW~\cite{dhall2011static}, and FER2013~\cite{FER2013}. In this section we briefly review the content of these databases.

\textbf{CMU MultiPIE:} CMU MultiPIE face database~\cite{gross2010multiPie} contains around 750,000 images of 337 people under multiple viewpoints, and different illumination conditions. There are four recording sessions in which subjects were instructed to display different facial expressions (i.e. Angry, Disgust, Happy, Neutral, Surprise, Squint, and Scream). We selected only the five frontal viewpoints (-45\degree to +45 \degree), giving us a total of around 200,000 images.

\textbf{MMI:} The MMI~\cite{pantic2005web} database includes more than 20 subjects of both genders (44\% female), ranging in age from 19 to 62, having either a European, Asian, or South American ethnicity. An image sequence is captured that has neutral faces at the beginning and the end for each session and subjects were instructed to display 79 series of facial expressions, six of which are prototypic emotions. We extracted static frames from each sequence, where it resulted in 11,500 images.


\textbf{CK+:} The Extended Cohn-Kanade database (CK+)~\cite{lucey2010extended} includes 593 video sequences recorded from 123 subjects ranging from 18 to 30 years old. Subjects displayed different expressions starting from the neutral for all sequences, and some sequences are labeled with basic expressions. We selected only the final frame of each sequence with peak expression in our experiment, which results in 309 images.

\textbf{DISFA:} Denver Intensity of Spontaneous Facial Actions (DISFA) database~\cite{mavadati2013disfa} is one of a few naturalistic databases that have been FACS coded with AU intensity values. This database consists of 27 subjects, each recorded while watching a four minutes video clip by two cameras. Twelve AUs are coded between 0-5, where 0 denotes the absence of the AU, while 5 represents maximum intensities. As DISFA is not emotion-specified coded, we used EMFACS system~\cite{friesen1983emfacs} to convert AU FACS codes to expressions, which resulted in around 89,000 images in which the majority have neutral expressions.

\textbf{FERA:} The GEMEP-FERA database~\cite{banziger2010introducing} is a subset of the GEMEP corpus used as database for the FERA 2011 challenge~\cite{valstar2011first}. It consists of recordings of 10 actors displaying a range of expressions. There are seven subjects in the training data, and six subjects in the test set. The training set contains 155 image sequences and the testing contains 134 image sequences. There are in total five emotion categories in the database: Anger, Fear, Happiness, Relief and Sadness. We extract static frames from the sequences with six basic expressions, which resulted to in around 7,000 images.

\textbf{SFEW:} The Static Facial Expressions in the Wild (SFEW) database~\cite{dhall2011static} is created by selecting static frames from Acted Facial Expressions in the Wild (AFEW)~\cite{dhall2013emotion}. The SFEW database covers unconstrained facial expressions, different head poses, age range, and occlusions and close to real world illuminations. There are a total of 95 subjects in the database. In total there are 663 well-labeled usable images.

\textbf{FER2013:} The Facial Expression Recognition 2013 (FER-2013) database was introduced in the ICML 2013 Challenges in Representation Learning~\cite{FER2013}. The database was created using the Google image search API and faces have been automatically registered. Faces are labeled as any of the six basic expressions as well as the neutral. The resulting database contains 35,887 images most of them in wild settings.

Table~\ref{tbl_databaseStatistic} shows the number of images for six basic expressions and neutral faces in each database.

\begin{table}[h]
\caption{Number of images per each expression in databases}
\label{tbl_databaseStatistic}
\centering
\begin{tabular}{l|c|c|c|c|c|c|c|}
\cline{2-8}
\multicolumn{1}{c|}{}                   & \scriptsize{\textbf{AN}} & \scriptsize{\textbf{DI}} & \scriptsize{\textbf{FE}} & \scriptsize{\textbf{HA}} & \scriptsize{\textbf{NE}} & \scriptsize{\textbf{SA}} & \scriptsize{\textbf{SU}} \\ \hline
\multicolumn{1}{|l|}{\scriptsize{\textbf{MultiPie}}} & \scriptsize{0}        & \scriptsize{22696}        & \scriptsize{0}        & \scriptsize{47338}        & \scriptsize{114305}        & \scriptsize{0}        & \scriptsize{19817}              \\ \hline
\multicolumn{1}{|l|}{\scriptsize{\textbf{MMI}}}      & \scriptsize{1959}        & \scriptsize{1517}        & \scriptsize{1313}        & \scriptsize{2785}        & \scriptsize{0}        & \scriptsize{2169}        & \scriptsize{1746}              \\ \hline
\multicolumn{1}{|l|}{\scriptsize{\textbf{CK+}}}      & \scriptsize{45}        & \scriptsize{59}        & \scriptsize{25}        & \scriptsize{69}        & \scriptsize{0}        & \scriptsize{28}        & \scriptsize{83}              \\ \hline
\multicolumn{1}{|l|}{\scriptsize{\textbf{DISFA}}}    & \scriptsize{436}        & \scriptsize{5326}        & \scriptsize{4073}        & \scriptsize{28404}        & \scriptsize{48582}        & \scriptsize{1024}        & \scriptsize{1365}              \\ \hline
\multicolumn{1}{|l|}{\scriptsize{\textbf{FERA}}}     & \scriptsize{1681}        & \scriptsize{0}        & \scriptsize{1467}        & \scriptsize{1882}        & \scriptsize{0}        & \scriptsize{2115}        & \scriptsize{0}              \\ \hline
\multicolumn{1}{|l|}{\scriptsize{\textbf{SFEW}}}     & \scriptsize{104}        & \scriptsize{81}        & \scriptsize{90}        & \scriptsize{112}        & \scriptsize{98}        & \scriptsize{92}        & \scriptsize{86}              \\ \hline
\multicolumn{1}{|l|}{\scriptsize{\textbf{FER2013}}}     & \scriptsize{4953}        & \scriptsize{547}        & \scriptsize{5121}        & \scriptsize{8989}        & \scriptsize{6198}        & \scriptsize{6077}        & \scriptsize{4002}              \\ \hline
\end{tabular}
    \begin{tablenotes}
      \footnotesize
      \item \textsuperscript{*} AN,	DI,	FE,	HA,	Ne,	SA,	SU stand for Anger, Disgust, Fear, Happiness, Neutral, Sadness, Surprised respectively.
    \end{tablenotes}
\end{table}

\subsection{Results}

We evaluated the accuracy of the proposed deep neural network architecture in two different experiments; viz. subject-independent and cross-database evaluation. In the subject-independent experiment, databases are split into training, validation, and test sets in a strict subject independent manner. We used the K-fold cross validation technique with K=5 to evaluate the results. In FERA and SFEW, the training and test sets are defined in the database release, and the results are evaluated on the database defined test set without performing K-fold cross validation. Since there are different samples per emotion per subject in some databases, the training, validation and test sets have slightly different sample sizes in each fold. On average we used 175K samples for training, 56K samples for validation, and 64K samples for test. The proposed architecture was trained for 200 epochs (i.e. 150K iterations on mini-batches of size 250 samples). Table~\ref{tbl_accuracy_PersonIndepndat} gives the average accuracy when classifying the images into the six basic expressions and the neutral expression. The average confusion matrix for subject-independent experiments can be seen in Table~\ref{tbl_Confusion_PersonIndepndat}.

Here, we also report the top-2 expression classes. As Table~\ref{tbl_accuracy_PersonIndepndat} depicts, the accuracy of the top-2 classification is 15\% higher than the top-1 accuracy in most cases, especially in the wild datasets (i.e. FERA, SFEW, FER2013). We believe that by assigning a single expression to a image can be ambiguous when there is transition between expressions or the given expression is not at its peak, and therefore the top-2 expression can result in a better classification performance when evaluating image sequences.

\begin{table}[h]
\caption{Average Accuracy (\%) for subject-independent}
\label{tbl_accuracy_PersonIndepndat}
\centering
\begin{tabular}{l|l|l|l|}
\cline{2-4}
                                                       & \multicolumn{1}{c|}{\textbf{\scriptsize{Top-1}}} & \multicolumn{1}{c|}{\textbf{\scriptsize{Top-2}}} & \multicolumn{1}{c|}{\textbf{\scriptsize{State-of-the-arts}}} \\ \hline
\multicolumn{1}{|l|}{\textbf{\scriptsize{MultiPIE}}} & \multicolumn{1}{c|}{\scriptsize{94.7$\pm$0.8}}            & \multicolumn{1}{c|}{\scriptsize{98.7$\pm$0.3}}            & \multicolumn{1}{c|}{\scriptsize{70.6~\cite{lee2014intra}, 90.6~\cite{eleftheriadis_TIP2014}}}            \\ \hline
\multicolumn{1}{|l|}{\textbf{\scriptsize{MMI}}}     & \multicolumn{1}{c|}{\scriptsize{77.6$\pm$2.9}}            & \multicolumn{1}{c|}{\scriptsize{86.8$\pm$6.2}}            & \multicolumn{1}{c|}{\scriptsize{63.4~\cite{liu2014deeply},  74.7~\cite{liu2013aware}, 79.8~\cite{mayer2014cross}, 86.9~\cite{shan2009facial}}}                   \\ \hline
\multicolumn{1}{|l|}{\textbf{\scriptsize{DISFA}}}    & \multicolumn{1}{c|}{\scriptsize{55.0$\pm$6.8}}            & \multicolumn{1}{c|}{\scriptsize{69.8$\pm$8.6}}            & \multicolumn{1}{c|}{\scriptsize{-}}             \\ \hline
\multicolumn{1}{|l|}{\textbf{\scriptsize{FERA}}}     & \multicolumn{1}{c|}{\scriptsize{76.7$\pm$3.6}}            & \multicolumn{1}{c|}{\scriptsize{90.5$\pm$4.6}}            & \multicolumn{1}{c|}{\scriptsize{56.1~\cite{liu2014deeply}, 75.0~\cite{FERAResults}, 55.6~\cite{valstar2011first}}}             \\ \hline
\multicolumn{1}{|l|}{\textbf{\scriptsize{SFEW}}}     & \multicolumn{1}{c|}{\scriptsize{47.7$\pm$1.7}}            & \multicolumn{1}{c|}{\scriptsize{62.1$\pm$1.2}}            & \multicolumn{1}{c|}{\scriptsize{26.1~\cite{liu2013aware}, 24.7~\cite{eleftheriadis_TIP2014}}}                      \\ \hline
\multicolumn{1}{|l|}{\textbf{\scriptsize{CK+}}}      & \multicolumn{1}{c|}{\scriptsize{93.2$\pm$1.4}}            & \multicolumn{1}{c|}{\scriptsize{97.8$\pm$1.3}}            & \multicolumn{1}{c|}{\begin{tabular}{@{}c@{}} \scriptsize{84.1~\cite{mayer2014cross}, 84.4~\cite{lee2014intra}, 88.5~\cite{taheri2014structure}, 92.0~\cite{liu2013aware}} \\ \scriptsize{92.4~\cite{liu2014deeply}, 93.6~\cite{zhang2015facial}}\end{tabular}}              \\ \hline
\multicolumn{1}{|l|}{\textbf{\scriptsize{FER2013}}}  & \multicolumn{1}{c|}{\scriptsize{66.4$\pm$0.6}}            & \multicolumn{1}{c|}{\scriptsize{81.7$\pm$0.3}}            & \multicolumn{1}{c|}{\scriptsize{69.3\cite{tang2013deep}}}               \\ \hline
\end{tabular}
\end{table}

\begin{table}[h]
\caption{Average (\%) confusion matrix for subject-independent}
\label{tbl_Confusion_PersonIndepndat}
\centering
\begin{tabular}{cc|c|c|c|c|c|c|c|}
\cline{3-9}
                                              &             & \multicolumn{7}{c|}{\scriptsize{predicted}}                                                                  \\ \cline{3-9}
                                              &             & \textbf{\scriptsize{AN}} & \textbf{\scriptsize{DI}} & \textbf{\scriptsize{FE}} & \textbf{\scriptsize{HA}} & \textbf{\scriptsize{NE}} & \textbf{\scriptsize{SA}} & \textbf{\scriptsize{SU}} \\ \hline
\multicolumn{1}{|c|}{\multirow{7}{*}{\begin{turn}{90}\scriptsize{Actual}\end{turn}
}} & \textbf{\scriptsize{AN}} & \scriptsize{\textbf{55.0}}        & \scriptsize{7.0}        & \scriptsize{12.8}        & \scriptsize{3.5}         & \scriptsize{7.6}        & \scriptsize{8.5}        & \scriptsize{5.3}         \\ \cline{2-9}
\multicolumn{1}{|c|}{}                        & \textbf{\scriptsize{DI}} & \scriptsize{1.0}        & \scriptsize{\textbf{80.3}}        & \scriptsize{1.8}         & \scriptsize{5.8}       & \scriptsize{8.5}        & \scriptsize{2.2}        & \scriptsize{0.1}         \\ \cline{2-9}
\multicolumn{1}{|c|}{}                        & \textbf{\scriptsize{FE}} & \scriptsize{7.4}         & \scriptsize{4.3}         & \scriptsize{\textbf{47.0}}        & \scriptsize{8.1}        & \scriptsize{18.7}        & \scriptsize{8.6}         & \scriptsize{5.5}        \\ \cline{2-9}
\multicolumn{1}{|c|}{}                        & \textbf{\scriptsize{HA}} & \scriptsize{0.7}         & \scriptsize{3.2}         & \scriptsize{2.4}         & \scriptsize{\textbf{86.6}}        & \scriptsize{5.5}         & \scriptsize{0.2}         & \scriptsize{1.0}         \\ \cline{2-9}
\multicolumn{1}{|c|}{}                        & \textbf{\scriptsize{NE}} & \scriptsize{2.3}         & \scriptsize{6.3}         & \scriptsize{7.8}         & \scriptsize{5.5}        & \scriptsize{\textbf{75.0}}        & \scriptsize{1.3}         & \scriptsize{1.4}         \\ \cline{2-9}
\multicolumn{1}{|c|}{}                        & \textbf{\scriptsize{SA}} & \scriptsize{6.0}           & \scriptsize{11.3}         & \scriptsize{8.9}        & \scriptsize{2.7}         & \scriptsize{13.7}        & \scriptsize{\textbf{56.1}}        & \scriptsize{0.9}         \\ \cline{2-9}
\multicolumn{1}{|c|}{}                        & \textbf{\scriptsize{SU}} & \scriptsize{0.8}         & \scriptsize{0.1}         & \scriptsize{2.8}        & \scriptsize{3.5}        & \scriptsize{2.5}        & \scriptsize{0.6}         & \scriptsize{\textbf{89.3}}        \\ \hline
\end{tabular}
\end{table}

\begin{figure}
\centering
\subfloat
{
	\centering
    \includegraphics[width=8cm]{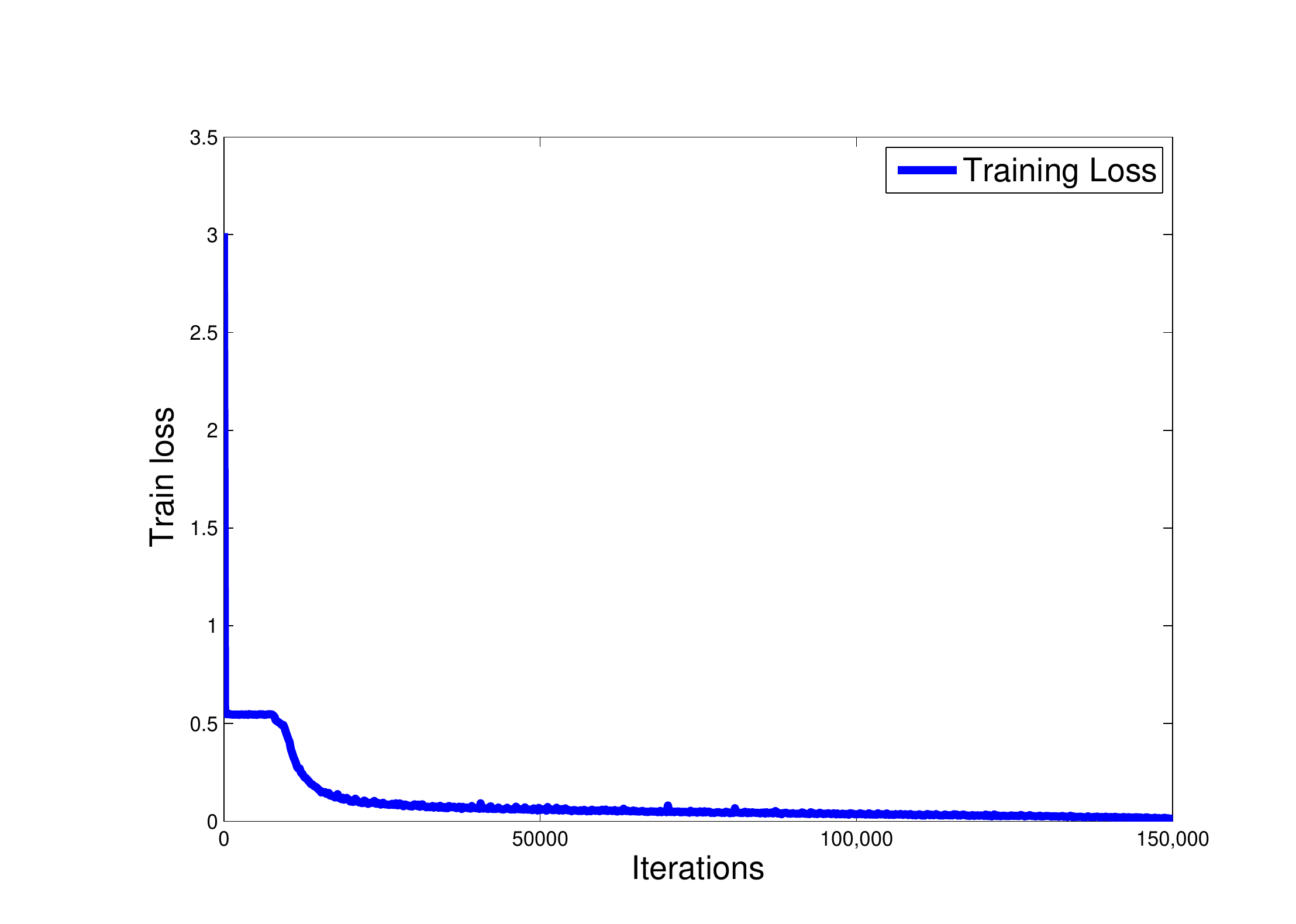}
}
\vspace{-0.6cm}
\subfloat
{
	\centering
    \includegraphics[width=8cm]{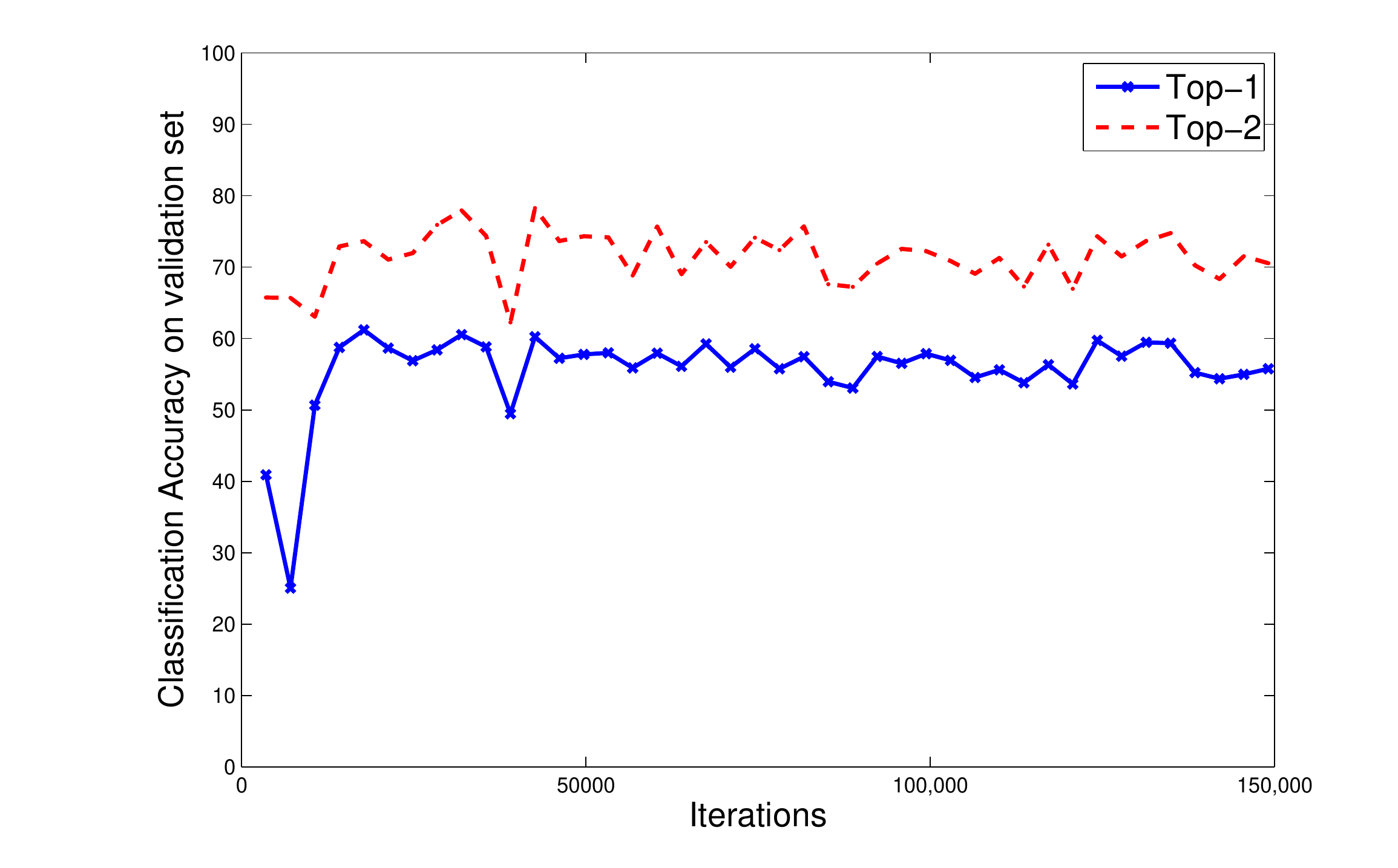}
}
\caption{\label{fig:training_TestingRate} Training loss and classification accuracy on validation set
}
\end{figure}

The proposed architecture was implemented using the Caffe toolbox~\cite{jia2014caffe} on a Tesla K40 GPU. It takes roughly 20 hours to train 175K samples for 200 epochs. Figure~\ref{fig:training_TestingRate} shows the training loss and classification accuracy of the top-1 and top-2 classification labels on the validation set of the subject-independent experiment over 150,000 iterations (about 150 epochs). As the figure illustrates, the proposed architecture converges after about 50 epochs.

In the cross-database experiment, one database is used for evaluation and the rest of databases are used to train the network. Because every database has a unique fingerprint (lighting, pose, emotions, etc.) the cross database task is much more difficult to extract features from (both for traditional SVM approaches, and for neural networks). The proposed architecture was trained for 100 epochs in each experiment. Table~\ref{tbl_accuracy_crossDB} gives the average cross-database accuracy when classifying the six basic expressions as well as the neutral expression.

\begin{table}[h]
\centering
\caption{Average Accuracy (\%) on cross database}
\label{tbl_accuracy_crossDB}
\centering
\begin{tabular}{l|l|l|l|l|l|l|}
\cline{2-7}
                                                       & \multicolumn{1}{c|}{\textbf{\scriptsize{top-1}}} & \multicolumn{1}{c|}{\textbf{\scriptsize{top-2}}} & \multicolumn{1}{c|}{\textbf{\scriptsize{\cite{mayer2014cross}}}} & \multicolumn{1}{c|}{\textbf{\scriptsize{\cite{shan2009facial}}}} & \multicolumn{1}{c|}{\textbf{\scriptsize{\cite{miao2012cross}}}} & \multicolumn{1}{c|}{\textbf{\scriptsize{\cite{zhang2015facial}}}} \\ \hline
\multicolumn{1}{|l|}{\textbf{\scriptsize{MultiPIE}}} & \multicolumn{1}{c|}{\scriptsize{45.7}}            & \multicolumn{1}{c|}{\scriptsize{63.2}}            & \multicolumn{1}{c|}{\scriptsize{-}}            & \multicolumn{1}{c|}{\scriptsize{-}}            & \multicolumn{1}{c|}{\scriptsize{-}}           & \multicolumn{1}{c|}{\scriptsize{-}} \\ \hline
\multicolumn{1}{|l|}{\textbf{\scriptsize{MMI}}}      & \multicolumn{1}{c|}{\scriptsize{55.6}}            & \multicolumn{1}{c|}{\scriptsize{68.3}}            & \multicolumn{1}{c|}{\scriptsize{51.4}}            & \multicolumn{1}{c|}{\scriptsize{50.8}}            & \multicolumn{1}{c|}{\scriptsize{36.8}}           & \multicolumn{1}{c|}{\scriptsize{66.9}}  \\ \hline
\multicolumn{1}{|l|}{\textbf{\scriptsize{DISFA}}}   & \multicolumn{1}{c|}{\scriptsize{37.7}}            & \multicolumn{1}{c|}{\scriptsize{53.2}}            & \multicolumn{1}{c|}{\scriptsize{-}}            & \multicolumn{1}{c|}{\scriptsize{-}}            & \multicolumn{1}{c|}{\scriptsize{-}}           & \multicolumn{1}{c|}{\scriptsize{-}} \\ \hline
\multicolumn{1}{|l|}{\textbf{\scriptsize{FERA}}}     & \multicolumn{1}{c|}{\scriptsize{39.4}}            & \multicolumn{1}{c|}{\scriptsize{58.7}}            & \multicolumn{1}{c|}{\scriptsize{-}}            & \multicolumn{1}{c|}{\scriptsize{-}}            & \multicolumn{1}{c|}{\scriptsize{-}}           & \multicolumn{1}{c|}{\scriptsize{-}}  \\ \hline
\multicolumn{1}{|l|}{\textbf{\scriptsize{SFEW}}}  & \multicolumn{1}{c|}{\scriptsize{39.8}}            & \multicolumn{1}{c|}{\scriptsize{55.3}}            & \multicolumn{1}{c|}{\scriptsize{-}}            & \multicolumn{1}{c|}{\scriptsize{-}}            & \multicolumn{1}{c|}{\scriptsize{-}}           & \multicolumn{1}{c|}{\scriptsize{-}}  \\ \hline
\multicolumn{1}{|l|}{\textbf{\scriptsize{CK+}}}      & \multicolumn{1}{c|}{\scriptsize{64.2}}            & \multicolumn{1}{c|}{\scriptsize{83.1}}            & \multicolumn{1}{c|}{\scriptsize{47.1}}            & \multicolumn{1}{c|}{\scriptsize{-}}            & \multicolumn{1}{c|}{\scriptsize{56.0}}           & \multicolumn{1}{c|}{\scriptsize{61.2}}  \\ \hline
\multicolumn{1}{|l|}{\textbf{\scriptsize{FER2013}}}  & \multicolumn{1}{c|}{\scriptsize{34.0}}            & \multicolumn{1}{c|}{\scriptsize{51.7}}            & \multicolumn{1}{c|}{\scriptsize{-}}            & \multicolumn{1}{c|}{\scriptsize{-}}            & \multicolumn{1}{c|}{\scriptsize{-}}           & \multicolumn{1}{c|}{\scriptsize{-}}  \\ \hline
\end{tabular}

\end{table}

The experiment presented in \cite{mayer2014cross} is a cross-database experiment performed by training the model on one of the CK+, MMI or FEEDTUM databases and testing the model on the others. The reported result in Table~\ref{tbl_accuracy_crossDB} is the average results for the CK+ and MMI databases.

Different classifiers on several databases are presented in~\cite{shan2009facial} where the results is still one of the state-of-the-art methods for person-independent evaluation on the MMI database (See Table~\ref{tbl_accuracy_PersonIndepndat}). The reported result in Table~\ref{tbl_accuracy_crossDB} is the best result using different SVM kernels trained on the CK+ database and evaluated the model on the MMI database.

A supervised kernel mean matching is presented in~\cite{miao2012cross} which attempts to match the distribution of the training data in a class-to-class manner. Extensive experiments were performed using four classifiers (SVM, Nearest Mean Classifier, Weighted Template Matching, and K-nearest neighbors). The reported result in Table~\ref{tbl_accuracy_crossDB} are the best results of the four classifier when training the model on the MMI and Jaffe databases and evaluating on the CK+ database as well as when the model was trained on the CK+ database and evaluated on the MMI database.

In~\cite{zhang2015facial} multiple features are fused via a Multiple Kernel Learning algorithm and the cross-database experiment is trained on CK+, evaluated on MMI and vice versa. Comparing the result of our proposed approach with these state-of-the-art methods, it can be concluded that our network can generalized well for FER problem. Unfortunately, there is not any study on cross-database evaluation of more challenging datasets such as FERA, SFEW and FER2013. We believe that this work can be a baseline for cross-database of these challenging datasets.

In \cite{eleftheriadis_TIP2014}, a Shared Gaussian Processes for multiview and viewinvariant classification is proposed. The reported result is very promising on MultiPIE database which covers multiple views, however on wild setting of SFEW it is not as efficient as MultiPIE. A new sparse representation is employed in~\cite{lee2014intra}, aiming to reduce the intra-class variation and by generating an intra-class variation image of each expression by using training images.

\cite{valstar2011first} is the FERA 2011 challenge baseline and \cite{FERAResults} is the result of UC Riverside team (winner of the challenge). \cite{taheri2014structure} detects AUs and uses their composition rules to recognize expressions by means of a dictionary-based approach, which is one of the state-of-the-art ''sign-based" approaches. \cite{tang2013deep} is the winner of the ICML 2013 Challenges on FER2013 database that employed a convolutional neural network similar to AlexNet~\cite{krizhevsky2012imagenet} but with linear one-vs-all linear SVM top layer instead of a Softmax function.


\begin{table}[h]
\caption{Subject-independent comparison with AlexNet results (\% accuracy)}
\label{tbl_benchmark_comparison}
\centering
\begin{tabular}{c|c|c|}
\cline{2-3}
 & \textbf{\scriptsize{Proposed Architecture}} & \textbf{\scriptsize{AlexNet}} \\
\cline{1-3}
\multicolumn{1}{|l|}{\textbf{\scriptsize{MultiPie}}} & \scriptsize{94.7} & \scriptsize{94.8} \\
\cline{1-3}
\multicolumn{1}{|l|}{\textbf{\scriptsize{MMI}}} & \scriptsize{77.9} & \scriptsize{56.0} \\
\cline{1-3}
\multicolumn{1}{|l|}{\textbf{\scriptsize{DISFA}}} & \scriptsize{55.0} & \scriptsize{56.1} \\
\cline{1-3}
\multicolumn{1}{|l|}{\textbf{\scriptsize{FERA}}} & \scriptsize{76.7} & \scriptsize{77.4} \\
\cline{1-3}
\multicolumn{1}{|l|}{\textbf{\scriptsize{SFEW}}} & \scriptsize{47.7} & \scriptsize{48.6} \\
\cline{1-3}
\multicolumn{1}{|l|}{\textbf{\scriptsize{CK+}}} & \scriptsize{93.2} & \scriptsize{92.2} \\
\cline{1-3}
\multicolumn{1}{|l|}{\textbf{\scriptsize{FER2013}}} & \scriptsize{66.4} & \scriptsize{61.1} \\
\cline{1-3}
\end{tabular}
\end{table}

As a benchmark to our proposed solution, we trained a full AlexNet from scratch (as opposed to fine tuning an already trained network) using the same protocol as used to train our own network. As shown in Table~\ref{tbl_benchmark_comparison}, our proposed architecture has better performance on MMI \& FER2013 and comparable performance on the rest of the databases. The value of the proposed solution over the AlexNet architecture is its training time - Our version of AlexNet performed more than 100M operations, whereas the proposed network performs about 25M operations.
\section{Discussion}

As shown in Tables~\ref{tbl_accuracy_PersonIndepndat} and~\ref{tbl_accuracy_crossDB}, the results in the subject-independent tests were either comparable to or better than the current state of the art. It should be mentioned that we have compared our results with the best methods on each database separately, where the hyper parameters of the presented models are fine-tuned for that specific problem. We perform significantly better than the state of the art on MultiPIE and SFEW (no known state of the art has been reported for the DISFA database). The only exceptions to the improved performance are with the MMI and FERA databases. There are a number of explanations for this phenomenon.


One of the likely reasons for the performance discrepancies on the subject-independent databases is due to the way that the networks are trained in our experiments. Because we use data from all of the studied databases to train the deep architecture, the input data contains image that do not conform to the database setting such as pose and lighting. It is very difficult to avoid this issue as it is hard or impossible to train such a complex network architecture on so little data without causing significant overfitting.
Another reason for the decreased performance is the focus on cross-database performance. By training slightly less complicated architectures, or even using traditional methods such as support vector machines, or engineered features, it would likely be possible to improve the performance of the network on subject-independent tasks. In this research however, we present a comprehensive solution that can generalize well to the FER ``in the wild'' problem. 



\section{Conclusion}
This work presents a new deep neural network architecture for automated facial expression recognition. The proposed network consists of two convolutional layers each followed by max pooling and then four Inception layers. The Inception layers increase the depth and width of the network while keeping the computational budget constant. The proposed approach is a single component architecture that takes registered facial images as the input and classifies them into either of the six basic expressions or the neutral.

We evaluated our proposed architecture in both subject-independent and cross-database manners on seven well-known publicly available databases. Our results confirm the superiority of our network compared to several state-of-the-art methods in which engineered features and classifier parameters are usually tuned on a very few databases. Our network is first which applies the Inception layer architecture to the FER problem across multiple databases. The clear advantage of the proposed method over conventional CNN methods (i.e. shallower or thinner networks) is gaining increased classification accuracy on both the subject independent and cross-database evaluation scenarios while reducing the number of operations required to train the network.

\section{Acknowledgment}
We gratefully acknowledge the support of NVIDIA Corporation with the donation of the Tesla K40 GPU used for this research. 

{\small
\bibliographystyle{ieee}

}

\end{document}